\documentclass{svproc}
\usepackage{amssymb}
\usepackage{enumitem}
\usepackage{amsmath}
\usepackage{booktabs}
\usepackage{multirow}
\usepackage[font=small,labelfont=bf]{caption}
\DeclareCaptionFont{tiny}{\tiny}
\usepackage{caption}
\usepackage{graphicx}
\usepackage{subcaption}
\begin{document}

\mainmatter

\title{An efficient density-based clustering algorithm using reverse nearest neighbour \thanks{This is an accepted manuscript to be presented at the Computing Conference 2019 in London. \textcopyright 2018. This manuscript version is made available under the CC-BY-NC-ND 4.0 license http://creativecommons.org/licenses/by-nc-nd/4.0/}}
\titlerunning{Density-based clustering}

\author{Stiphen Chowdhury\inst{1} \and Renato Cordeiro de Amorim\inst{2}}
\authorrunning{S. Chowdhury \and R.C. de Amorim} 
\tocauthor{Stiphen Chowdhury, Renato Cordeiro de Amorim}

\institute{University of Hertfordshire, School of Computer Science, College Lane Campus, Hatfield AL10 9AB, UK,\\
\email{s.chowdhury8@herts.ac.uk},\\ 
\and
University of Essex,
School of Computer Science and Electronic Engineering, Wivenhoe Park, Colchester CO4 3SQ\\
\email{r.amorim@essex.ac.uk}}

\maketitle 

\begin{abstract}
Density-based clustering is the task of discovering high-density regions of entities (clusters) that are separated from each other by contiguous regions of low-density. DBSCAN is, arguably, the most popular density-based clustering algorithm. However, its cluster recovery capabilities depend on the combination of the two parameters. In this paper we present a new density-based clustering algorithm which uses reverse nearest neighbour (RNN) and has a single parameter. We also show that it is possible to estimate a good value for this parameter using a clustering validity index. The RNN queries enable our algorithm to estimate densities taking more than a single entity into account, and to recover clusters that are not well-separated or have different densities. Our experiments on synthetic and real-world data sets show our proposed algorithm outperforms DBSCAN and its recent variant ISDBSCAN.
\end{abstract}

\section{Introduction}
\label{Sec:Introduction}
Clustering algorithms aim to reveal natural groups of entities within a given data set. These groups (clusters) are formed in such a way that each contains homogeneous entities, according to a pre-defined similarity measure. This grouping of similar entities is usually data-driven and by consequence it does not require information regarding the class label of the entities. Detecting, analysing, and describing natural groups within a data set is of fundamental importance to a number of scientific fields. Thus, it is common to see clustering algorithms being applied to problems in various fields such as: bioinformatics, image processing, astronomy, pattern recognition, medicine, and marketing \cite{Jain:1999:DCR:331499.331504,7460951,JAIN2010651,mirkin2012clustering}.

There are indeed a number of different approaches to clustering. Some algorithms were designed so they could be applied to data sets in which each entity is described over a number of features. Others, take as input a dissimilarity matrix or even the weights of edges in a graph. There are different formats for the final clustering as well. The clusters may be a partition of the original data set, or they may present overlaps so that an entity belongs to more than one cluster (usually at different degrees, adding to one). They may also be non-exhaustive so that not every entity belongs to a cluster, which can be particularly helpful if the data set contains noise entities. We may also have hierarchical clusterings, which may be generated following a top-down or bottom-up approach. We direct readers interested in more details to the literature (see for instance \cite{Jain:1999:DCR:331499.331504,JAIN2010651,mirkin2012clustering} and references therein). In this paper we focus on density-based clustering. This approach defines clusters as areas of higher density separated by areas of lower density. Clearly, such loose definition may raise a number of questions regarding what exactly a cluster is (or is not!). However, given there is no generally accepted definition for the term \textit{cluster} that works in all scenarios, one can raise similar questions even if using non density-based algorithms. Defining `true' clusters is particularly difficult and may also depend on other factors than the data set alone (for a discussion see \cite{hennig2015true} and references therein). The major advantage a density-based algorithm has is that the impact of a similarity measure on the shape bias of clusters is considerably reduced.

Density-Based Spatial Clustering of Applications with Noise (DBSCAN) \cite{Ester_1996} is arguably the most popular density-based clustering algorithm. A recent search in Google Scholar for the term ``DBSCAN'' returned a total of 21,500 entries. Most importantly searches for the years of 2014, 2015, and 2016 returned 2,190, 2,710, and 3,550, respectively. These numbers support the growing popularity of DBSCAN. Unfortunately, as popular as it may be, DBSCAN is not without weaknesses. For instance: (i) it requires two parameters (for details see Section \ref{Sec:RelatedWork}); (ii) it is a non-deterministic algorithm, so it may produce different partitions under the same settings; (iii) it is not particularly suitable for data sets whose clusters have different densities.

There have been some advancements in the literature. For instance, OPTICS \cite{ankerst1999optics} has been designed to deal with clusters of different densities. Using the concept of $k$-Influence Space \cite{hinneburg1998efficient,jin2006ranking}, ISDBSCAN \cite{CASSISI2013317} can also deal with clusters of different densities, and requires a single parameter to be tuned. ISDBSCAN algorithm significantly outperforms DBSCAN and OPTICS \cite{ankerst1999optics}.

In this paper we make a further advancement in density-based clustering research. Here, we introduce Density-based spatial clustering using reverse nearest neighbour (DBSCRN) a new method capable of matching or improving cluster recovery in comparison to ISDBSCAN (and by consequence DBSCAN and OPTICS), but being orders of magnitude faster. Our method has a single parameter for which we show a clear estimating method.

\section{Related work}
\label{Sec:RelatedWork}

The purpose of any clustering algorithms is to split a data set $Y$ containing $n$ entities $y_i \in \mathbb{R}^m$ into $K$ clusters $S=\{S_1, S_2, ..., S_K\}$. Here, we are particularly interested in hard-clustering so that a given entity $y_i$ can be assigned to a single cluster $S_k \in S$. Thus, 
the final clustering is a partition subject to $S_k \cap S_l = \emptyset$ for $k,l = 1, 2, ..., K$ and $k\neq l$. 

It is often stated that density-based clustering algorithms are capable of recovering clusters of arbitrary shapes. This is a very tempting thought, which may lead to some disregarding the importance of selecting an appropriate distance or similarity measure. This measure is the key to produce homogeneous clusters as it defines homogeneity. Selecting a measure will have an impact on the actual clustering. Most likely the impact will not be as obvious as if one were to apply an algorithm such as $k$-means \cite{macqueen1967some} (where the measure in use leads to a bias towards a particular cluster shape). However, the impact of this selection will still exist at a more local level. If this was not the case, DBSCAN would produce the same clustering regardless of the distance measure in place.

Arguably, the most popular way of calculating the dissimilarity between two entities $y_i, y_j$ each described over $m$ features is given by the squared Euclidean distance, that is
\begin{equation}
\label{Eq:EuclideanDistance}
d(y_i, y_j) = \sum_{v=1}^m (y_{iv}-y_{jv})^2.
\end{equation}
DBSCAN classifies each entity $y_i \in Y$ as either a core entity, a reachable entity, or an outlier. To do so, this algorithm applies (\ref{Eq:EuclideanDistance}) together with two parameters: a distance threshold ($\epsilon$), and the minimum number of entities required to form a dense region ($MinPts$). The $\epsilon-$neighbourhood of an entity $y_i \in Y$ is given by
\begin{equation}
\label{Eq:EpsilonNeighborhood}
\mathcal{N}(y_i) = \{y_j \in Y \mid d(y_i, y_j) \leq \epsilon\},
\end{equation}
so that $\mathcal{N}(y_i) \subseteq Y$. Clearly, $\mathcal{N}(y_i) = Y$ would be an indication the value of $\epsilon$ is too high. An entity $y_i \in Y$ is classified as a core entity iff
\begin{equation}
|\mathcal{N}(y_i)|\geq MinPts,
\end{equation}
in this case each entity in $\mathcal{N}(y_i)$ is said to be directly reachable from $y_i$. No entity can be directly reachable from a non-core entity. An entity $y_i$ is classified as a reachable entity if there is a path  $y_j, y_{j+1}, y_{j+2}, ..., y_i$ in which each entity is directly reachable from the previous. If these two cases (core and reachable) do not apply, then $y_i$ is classified as an outlier. Given a core entity $y_i$, DBSCAN can form a cluster of entities (core and non-core) that are reachable from $y_i$. The general idea is, of course, very intuitive but one may find difficult to set $\epsilon$ and $MinPts$ as they are problem-dependent.

The ISDBSCAN outperforms the above and OPTICS. Probably, the major reason for this is the use of the $k$-influence space ($IS_k$) to define the density around a particular entity. $IS_k$ is based on the $k$-nearest neighbour ($NN_k$) \cite{altman1992introduction} and reverse $k$-nearest neighbour ($RNN_k$) \cite{Korn:2000:ISB:335191.335415} methods. 
\begin{equation}
\label{Eq:NearestNeighbor}
	NN_k(y_i) = \{y_1, y_2,..., y_j, ..., y_k \in Y \mid d(y_j, y_i) \leq d(y_t, y_i) \forall y_t \in Y^{\prime}\},
\end{equation}
where $Y^{\prime} = Y \backslash \{y_1, y_2, ..., y_j, ..., y_k\}$, and $k$ is the number of nearest neighbours. The reverse $k$-nearest neighbours  is given by the set

\begin{equation}
\label{Eq:ReverseNearestNeighbor}
	RNN_k(y_i) = \{y_j \in Y \mid y_i \in NN_k(y_j)\},
\end{equation}
leading to the $k-$influence space
\begin{equation}
	IS_k(y_i) = NN_k(y_i) \cap RNN_k(y_i).
\end{equation}
With the above we can now describe ISDBSCAN.
\\\\ \textbf{ISDBSCAN}$(Y,k)$\\
\textit{Input}\\
$Y$: Data set to be clustered;\\
$k$: Number of nearest neighbours;\\
\textit{Output}\\
$S$: A clustering $S=\{S_1, S_2, ...,S_c, ..., S_K\}$;\\
$S_{noise}$: A set of entities marked as noise;\\
\textit{Algorithm:}
\begin{enumerate}[nolistsep]
\item while $Y \neq \emptyset$
\item \hspace{4mm} Randomly select $y_i$ fom $Y$;
\item \hspace{4mm} $S_c \leftarrow$ MakeCluster($Y$,$y_i$,$k$);
\item \hspace{4mm} $Y \leftarrow Y \backslash S_c$;
\item \hspace{4mm} if $|S_c| > k$ then
\item \hspace{8mm} Add $S_c$ to S;
\item \hspace{4mm} else
\item \hspace{8mm} Add $y_i$ to $S_{noise}$;
\item \hspace{4mm} end if
\item end while
\item return $S$; \\ 
\item \textit{MakeCluster($Y$,$y_i$,$k$)}
\item $S_c \leftarrow \emptyset$;
\item if $|IS_k(y_i)| > 2 / 3k$ then
\item \hspace{4mm} for each $y_j \in IS_k(y_i)$ do
\item \hspace{8mm} $S_c \leftarrow S_c \cup \{y_j\}$;
\item \hspace{8mm} $S_c= S_c \cup $ MakeCluster($Y$,$y_j$,$k$);
\item \hspace{4mm} end for
\item endif
\item return $S_c$ \\
\end{enumerate}

\section{Density-based spatial clustering using reverse nearest neighbour (DBSCRN)}

The algorithm we introduce in this paper, DBSCRN, has some similarities to DBSCAN. They are both density-based clustering algorithms which need to determine whether an entity $y_i \in Y$ is core or non-core. Section \ref{Sec:RelatedWork} explains how this is done by DBSCAN. In the case of DBSCRN this is determined using a reverse nearest neighbour query. Given an entity $y_i \in Y$ we apply Equation (\ref{Eq:ReverseNearestNeighbor}) to find the set of 
entities to which $y_i$ is one of their $k$-nearest neighbours. We find this to be a more robust method to estimate density because it uses more than just one core entity to find nearest neighbours. We present the DBSCRN algorithm below.
\\\\ \textbf{\textbf{DBSCRN}}$(Y, k)$\\
\textit{Input}\\
$Y$: Data set to be clustered.\\
$k$: Number of nearest neighbours.\\ 
\textit{Output}\\
$S:$ A clustering $S = \{S_1,S_2, \cdots, S_K \}$ \\
\textit{Algorithm:}
\begin{enumerate}[nolistsep]
\item for each $y_i \in Y$ do
\item \hspace{4mm} if $|RNN_{k}(y_i)|$ $ < $ $k$ then
\item \hspace{8mm} Add $y_i$ to $S_{non-core}$;
\item \hspace{4mm} else
\item \hspace{8mm} Add $y_i$ to $S_{core}$;
\item \hspace{8mm} $S \longleftarrow S \cup $  \textit{expandCluster($y_i$,$k$, $S$)};
\item \hspace{4mm} end if
\item end for
\item Assign each $y_j \in S_{non-core}$ to the cluster of the nearest $y_i \in S_{core}$, using Equation (\ref{Eq:EuclideanDistance});
\item return $S$; \\ 

\item \textit{expandCluster($y_i$,$k$, $S$)}
\item $S_{y_i} \leftarrow \{y_i\}$;
\item $S_{tmp} \leftarrow \{y_i\}$;
\item for each $y_j \in RNN_{k}(y_k \in S_{tmp})$ do
\item \hspace{4mm} if $|RNN_{k}(y_j)|$ $ > $ $2k/\pi$ then
\item \hspace{8mm} $S_{tmp} \leftarrow S_{tmp} \cup RNN_{k}(y_j)$;
\item \hspace{4mm} end if
\item \hspace{4mm} If $y_j \notin S_{tmp}$ and $y_j$ is not assigned to any cluster in $S$.
\item \hspace{8mm} Add $y_j$ to $S_{y_i}$;
\item \hspace{4mm} end if
\item end for
\item return $S_{y_i}$; \\
\end{enumerate}
In the above the quantity of nearest neighbours ($k$) is a user-defined parameter. The quantity of clusters ($K$) is automatically found by the algorithm.

%
\section{Estimating parameters}

Here, we take the view that parameter estimation can be accomplished using a Clustering Validity Index (CVI). Validation is one of the most challenging aspects of clustering. It raises the question: how can one measure the quality of a clustering when labeled data is non-existent? the simple fact an algorithm produced a clustering says nothing about the quality of that clustering. Clustering algorithms will produce a clustering even if the data has no cluster structure. A number of CVIs have been proposed to measure the quality of clusterings obtained using distance-based algorithms such as \textit{k}-means (for a review see \cite{arbelaitz2013extensive} and references therein). Selecting a CVI to use is not a trivial matter, it should take into account the definition of cluster in use and any other requirement that may exist. CVIs suitable for density-based clustering algorithms are not as popular. However, they are particularly important as all algorithms we experiment with have at least one parameter that needs to be estimated.

In this paper we do not focus on finding and comparing CVIs suitable for density-based clustering algorithms. One could apply any such CVI to estimate the parameters of the methods we experiment with. With this in mind we leave such comparison for future work. Here, we have experimented with Density-Based Clustering Validation (DBCV) \cite{MoulaviJCZS14}. This CVI measures clustering quality based on the relative density connection between pairs of entities. This index is formulated on the basis of a new kernel density function, which is used to compute the density of entities and to evaluate the within and between-cluster density connectedness of clustering results. This is well aligned to the definition we use of cluster (See section \ref{Sec:Introduction}).

Using density-based clustering algorithms, DBCV has unsurprisingly outperformed the Silhouette Width \cite{rousseeuw1987silhouettes}, the Variance Ratio Criterion \cite{calinski1974dendrite}, and Dunn's index \cite{dunn1974well}. These three CVIs are not well-aligned with the definition of cluster used by density-based clustering algorithms. DBCV has also outperformed Maulik-Bandyopadhyay \cite{maulik2002performance} and CDbw \cite{halkidi2008density}. 
\section{Setting of experiments}
We experimented with synthetic and real-world data sets, all obtained from the UCI machine learning repository \cite{bache2013uci}. We selected the data sets described in Table \ref{Tb:datasets}, these are rather popular and have been used in a number of publications \cite{fu2007flame,jain2005data,veenman2002maximum,chang2008robust,zahn1971graph,gionis2007clustering,AHG:AHG2137,tan1988using,fisher2014concept}. The clusters in real-world data sets, like Iris, tend to have a globular shape aligned to Gaussian distributions. The synthetic data sets contain arbitrarily shaped clusters of different sizes and densities. All of these data sets allow us to scrutinize the cluster recovery of the clustering algorithms we experiment with.
\begin{table}[htb]
\caption{Data sets used in our experiments.}
\centering
\begin{tabular}{lccc}
\toprule
& Entities & Clusters & Features \\ \midrule
Aggregation & 788   & 7   & 2 \\
Compound    & 399   & 6   & 2 \\
Pathbased   & 300   & 3   & 2 \\
Spiral      & 200   & 2   & 2 \\
Mixed       & 1479  & 5   & 2 \\
Toy         & 373   & 2   & 2 \\ 
Flame       & 240   & 2   & 2 \\ 
R15         & 600   & 15  & 2 \\
Soya        & 47    & 4   & 58 \\ 
Iris        & 150   & 3   & 4 \\ \hline \bottomrule 
\end{tabular}
\label{Tb:datasets}
\end{table} 

We have the set of correct labels for each of the data sets we experiment with. This allows us to measure the cluster recovery of each algorithm in relation to the correct labels. In each experiment we generate a set of labels from a clustering solution using a confusion matrix. We then compare the labels of the clustering solution with the correct labels using the adjusted Rand Index (ARI) \cite{hubert1985comparing}.
\begin{equation*} 
ARI=\frac{\sum\nolimits_{ij}\left( \begin{array}{c} {n_{ij}} \\ 2 \end{array} \right)-{\left[ \sum\nolimits_i\left( \begin{array}{c} {a_i} \\ 2 \end{array} \right)\sum\nolimits_j\left( \begin{array}{c} {b_j} \\ 2 \end{array} \right)\right]}/{\left( \begin{array}{c} {n} \\ 2 \end{array} \right)}}{\frac{1}{2}\left[ \sum\nolimits_i\left( \begin{array}{c} {a_i} \\ 2 \end{array} \right)\sum\nolimits_j\left( \begin{array}{c} {b_j} \\ 2 \end{array} \right)\right]-{\left[ \sum\nolimits_i\left( \begin{array}{c} {a_i} \\ 2 \end{array} \right)\sum\nolimits_j\left( \begin{array}{c} {b_j} \\ 2 \end{array} \right)\right]}/{\left( \begin{array}{c} {n} \\ 2 \end{array} \right)}},
\end{equation*}
where $n_{ij} = \left| S_i \cap S_j\right|$, $a_i=\sum\nolimits_{j=1}^K\left| S_i \cap S_j\right|$ and $b_i=\sum\nolimits_{i=1}^K\left| S_i \cap S_j\right|$.

We have standardised the features of each data set by their respective ranges
\begin{equation}
	\label{Eq:Stand}
	y_{iv} = \frac{y_{iv} - \bar{y_v}}{max(y_v) - min(y_v)},	
\end{equation}
where $\bar{y_v}= n^{-1} \sum_{i=1}^n y_{iv}$. We chose to use (\ref{Eq:Stand}) rather than the popular $z$-score because the latter favours unimodal distributions. For instance, consider two features: a unimodal $v_1$ and a bimodal $v_2$. The standard deviation of $v_2$ will be higher than that of $v_1$. By consequence the $z$-score (and the contribution to the clustering) of $v_2$ will be lower than that of $v_1$. However, we would be usually interested in the cluster structure present in $v_2$.

We experiment with three algorithms: DBSCAN, ISDBSCAN, and DBSCRN. Each of these algorithms require the use of parameters, we have estimated these using DBCV. In the case of DBSCAN we run experiments with values for $MinPts$ from 3 to 20 in steps of 1, and $\epsilon$ from the minimum pairwise distance to the maximum pairwise distance in steps of $0.1$. We selected as final clustering that with the best DBCV index. For ISDBSCAN, we run experiments setting the number of nearest neighbours from 5 to 25 in steps of 1. In the case of DBSCRN we experiment with values of $k$ (the number of nearest neighbours) from 3 to 30, in steps of 1.

All experiments were run on a PC with Intel(R) Core(TM) i7-2670QM CPU 2.20GHz and 8.00GB RAM. The operating system was Windows 7 (64-bits). The algorithms were implemented using MATLAB 2016a.

\section{Results and Discussion}

In this section we present the results, and discussion, of our experiments. We compare $k$-means, DBSCAN, ISDBSCAN, and DBSCRN on the data sets presented in Table \ref{Tb:datasets}. Our comparison is mainly focused on cluster recovery, measured using the ARI, but we also discuss the amount of time the algorithms take to complete.

In our first set of experiments we aim to show the best possible cluster recovery for each algorithm. Given an algorithm, we set its parameters to those producing a clustering with the highest ARI. This scenario is not realistic as it requires the user to know the correct labels for each data set. However, it allows us to analyse the best possible result for each algorithm. Table \ref{Tb:BestPossibleResults} presents the results for this set of experiments. Each non-deterministic algorithm was run 100 times.
\begin{table*}[t]
\caption{Experiments comparing $k$-means, DBSCAN, ISDBSCAN, and DBSCRN. This table reports the best possible ARI each of the algorithms can achieve at each data set. Non-deterministic algorithms were run 100 times.}
\centering
\begin{tabular}{llllllllllcl}
\toprule
\multirow{2}{*}{} & \multicolumn{3}{c}{$k$-means}  & \multicolumn{3}{c}{DBSCAN} & \multicolumn{3}{c}{ISDBSCAN} &\multicolumn{2}{c}{DBSCRN} \\ 
\cmidrule(lr){2-4} \cmidrule(lr){5-7} \cmidrule(lr){8-10} \cmidrule(lr){11-12}
  & Mean & Std dev & Max & Mean & Std dev & Max &Mean & Std dev & Max& Mean & Max  \\ \midrule
Aggregation &0.74&0.03&0.78&0.98&0.002&0.98&0.91  & 0.02   & 0.94  &  -  & \textbf{0.99} \\
Compound    & 0.57  & 0.10   & 0.78  &0.83  & 0.00   & 0.83   &0.88  & 0.01   & 0.91 &	-  & \textbf{0.96} \\
Pathbased   & 0.46  & 0.001   & 0.46  &0.89  & 0.01   & 0.9&0.85  & 0.01   & 0.89 &  -  & \textbf{0.92} \\
Spiral      & 0.05  & 0.01    & 0.06  &\textbf{1.00}  & 0.00   & 1.00&0.98  & 0.00   & 1.00 &  -  & \textbf{1.00} \\
Mixed       & 0.39  & 0.02    & 0.42  &\textbf{1.00}  & 0.00   & 1.00&\textbf{1.00}  & 0.00   & 1.00& 	-  & \textbf{1.00} \\
Toy         & 0.31  & 0.01    & 0.32  &0.96  & 0.00   & 0.96&\textbf{1.00}  & 0.00   & 1.00& 	-  & \textbf{1.00} \\ 
Flame       & 0.46  & 0.02    & 0.51  &\textbf{0.96}  & 0.00   &0.96  & 0.90   & 0.00&0.90& 	-  & 0.93 \\ 
R15         & 0.88  & 0.07    & 0.99  &\textbf{0.99}  & 0.00   & 0.99&0.94  & 0.00   & 0.94& 	-  & \textbf{0.99} \\
Soya        & 0.80  & 0.20    & 1.00  &\textbf{1.00}  & 0.00   & 1.00&0.96  & 0.00   & 0.96& 	-  & \textbf{1.00} \\ 
Iris        & \textbf{0.67}  & 0.10    & 0.71  &0.36  & 0.01   & 0.37& 0.40  & 0.01   & 0.47&	-  & 0.45 \\ \hline \bottomrule
\end{tabular}
\label{Tb:BestPossibleResults}
\end{table*}

Table \ref{Tb:BestPossibleResults} shows that in the vast majority of cases our method is competitive or superior to others in average. The noticeable exception is given by $k$-means in the Iris data set. In this case none of the density-based clustering algorithms performs well. Most likely, the definition of cluster used in $k$-means (a globular set of entities in the Euclidean space) is more well-aligned to the clusters in this particular data set. This should remind us that one should define what a cluster is before choosing a clustering algorithm.  

\begin{figure}[!htb]
    \centering
    \caption{Best possible cluster recovery as measured by the ARI on the Compound data set.}
    \label{Fig:Compound_NestPossible}
    \begin{subfigure}[b]{0.5\textwidth}
          \centering
          \begin{subfigure}[b]{0.5\textwidth}
                  \centering
                  \includegraphics[width=\textwidth]{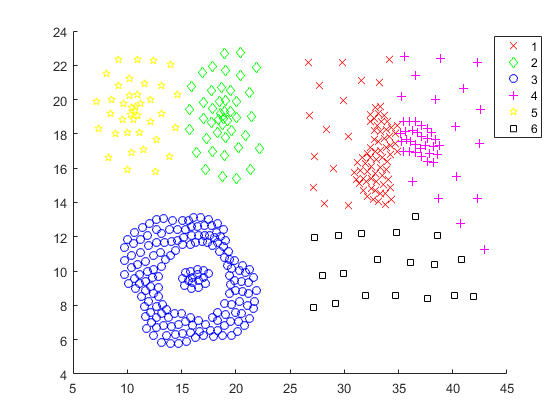}
                  \caption{$k$-means}
          \end{subfigure}%
          \begin{subfigure}[b]{0.5\textwidth}
                  \centering
                  \includegraphics[width=\textwidth]{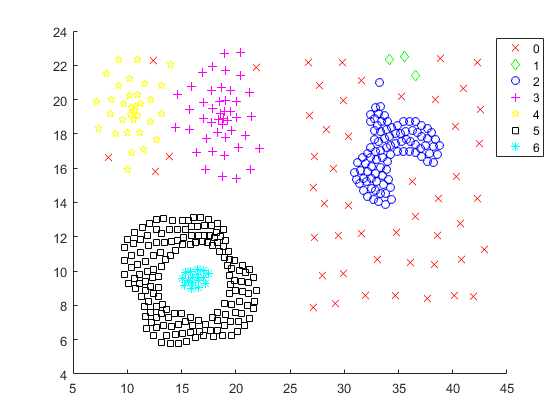}
                  \caption{DBSCAN}
          \end{subfigure}
                    \centering
          \begin{subfigure}[b]{0.5\textwidth}
                  \centering
                  \includegraphics[width=\textwidth]{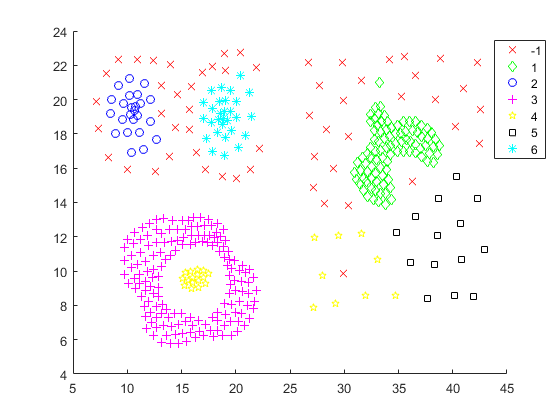}
                  \caption{ISDBSCAN}
          \end{subfigure}%
          \begin{subfigure}[b]{0.5\textwidth}
                  \centering
                  \includegraphics[width=\textwidth]{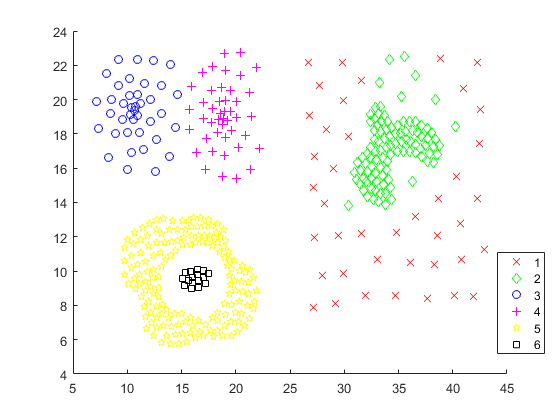}
                  \caption{DBSCRN}
          \end{subfigure}
    \end{subfigure}
\end{figure}

Let us analyse in more details some of the results in Table \ref{Tb:BestPossibleResults}. The Compound data set contains three difficult clustering problems: (i) nested clusters with approximately the same density; (ii) nested clusters with different densities; (iii) clusters separated by local minimum density regions. This data set contains two clusters for each of these problems. Figure \ref{Fig:Compound_NestPossible} presents the best possible results for each of the algorithms we experiment with. The $k$-means algorithm searches for globular clusters in the data set, so it is unable to deal with problems (i) and (ii). Probably the major weakness of DBSCAN is its inability to detect clusters of different densities, leading to 51 out of 399 entities being classifies as noise (red cross, labelled as zero). ISDBSCAN was designed to deal with clusters of different densities, but does not deal well with problems (ii) and (iii) on this occasion. Our method does produce misclassification, but there are considerably less of them than in other methods.

\begin{figure}[!htb]
    \centering
    \caption{Best possible cluster recovery as measured by the ARI on the Flame data set.}
    \label{Fig:Flame_BestPossible}
    \begin{subfigure}[b]{0.5\textwidth}
          \centering
          \begin{subfigure}[b]{0.5\textwidth}
                  \centering
                  \includegraphics[width=\textwidth]{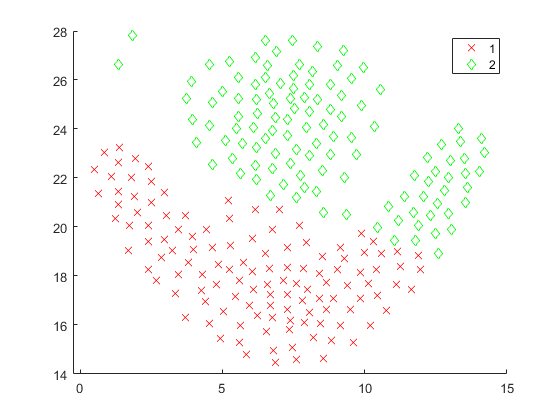}
                  \caption{$k$-means}
          \end{subfigure}%
          \begin{subfigure}[b]{0.5\textwidth}
                  \centering
                  \includegraphics[width=\textwidth]{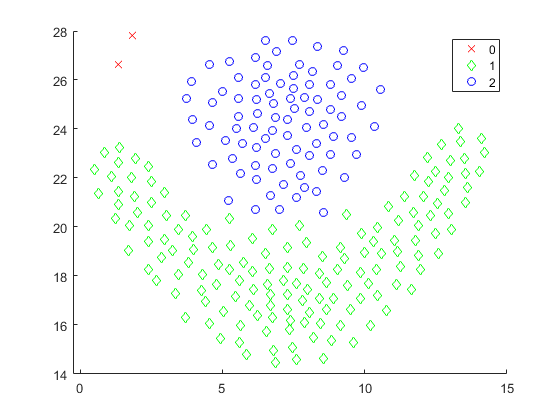}
                  \caption{DBSCAN}
          \end{subfigure}
                    \centering
          \begin{subfigure}[b]{0.5\textwidth}
                  \centering
                  \includegraphics[width=\textwidth]{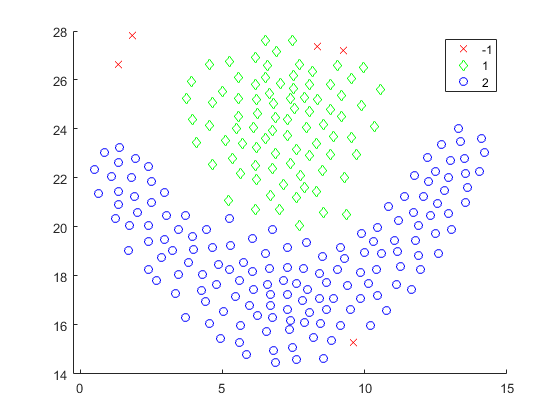}
                  \caption{ISDBSCAN}
          \end{subfigure}%
          \begin{subfigure}[b]{0.5\textwidth}
                  \centering
                  \includegraphics[width=\textwidth]{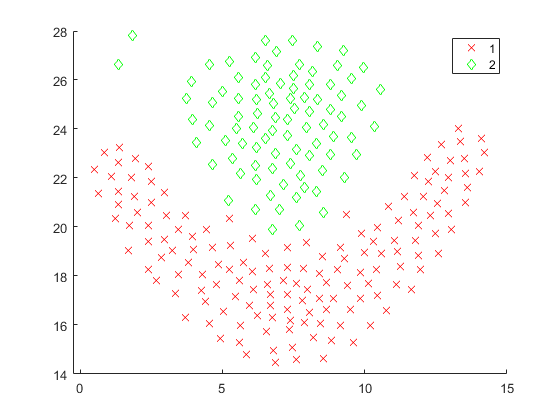}
                  \caption{DBSCRN}
          \end{subfigure}
    \end{subfigure}
\end{figure}

The Flame data set contains two clusters of similar densities separated by either a low density region or a soft boundary. Figure \ref{Fig:Flame_BestPossible} presents the best possible clusterings for each algorithm. We can see $k$-means is unable to correctly separate these clusters, as they are not Gaussian. DBSCAN does perform particularly well in this data set, but as well as ISDBSCAN it wrongly classifies a few entities as noise. In the case of ISDBSCAN this happens because there is a lower cluster density near the boundary region, leading to the misclassification of entities as noise.

\begin{figure}[!htb]
    \centering
    \caption{Best possible cluster recovery as measured by the ARI on the Pathbased data set.}
    \label{Fig:Pathbased_BestPossible}
    \begin{subfigure}[b]{0.5\textwidth}
          \centering
          \begin{subfigure}[b]{0.5\textwidth}
                  \centering
                  \includegraphics[width=\textwidth]{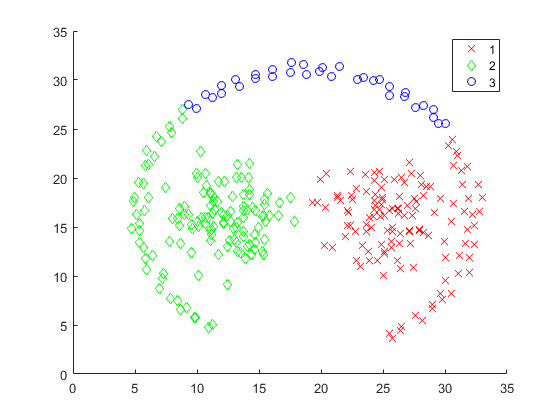}
                  \caption{$k$-means}
          \end{subfigure}%
          \begin{subfigure}[b]{0.5\textwidth}
                  \centering
                  \includegraphics[width=\textwidth]{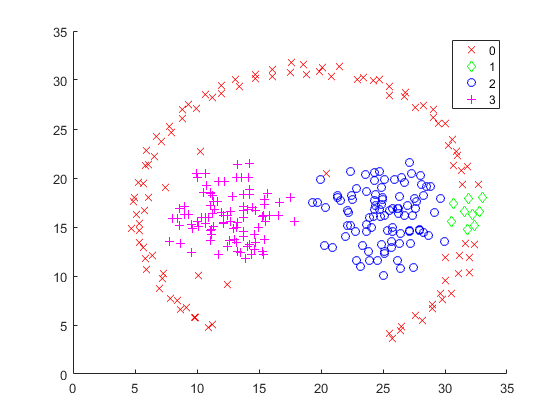}
                  \caption{DBSCAN}
          \end{subfigure}
                    \centering
          \begin{subfigure}[b]{0.5\textwidth}
                  \centering
                  \includegraphics[width=\textwidth]{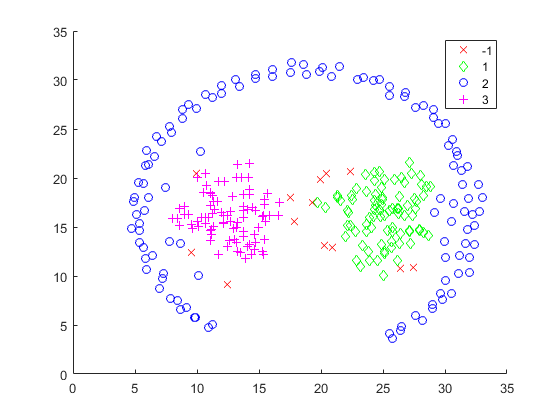}
                  \caption{ISDBSCAN}
          \end{subfigure}%
          \begin{subfigure}[b]{0.5\textwidth}
                  \centering
                  \includegraphics[width=\textwidth]{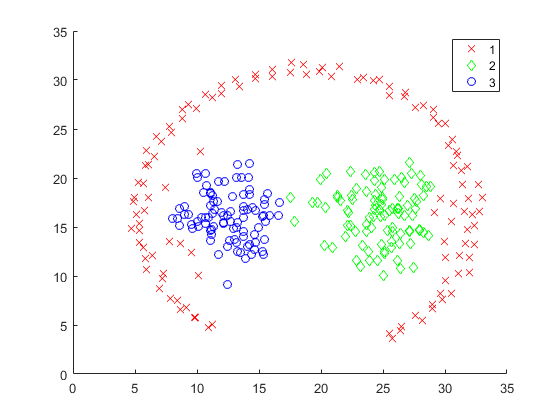}
                  \caption{DBSCRN}
          \end{subfigure}
    \end{subfigure}
\end{figure}

\begin{figure}[!htb]
    \centering
    \caption{Best possible cluster recovery as measured by the ARI on the Toy data set.}
    \label{Fig:Toy_BestPossible}
    \begin{subfigure}[b]{0.5\textwidth}
          \centering
          \begin{subfigure}[b]{0.5\textwidth}
                  \centering
                  \includegraphics[width=\textwidth]{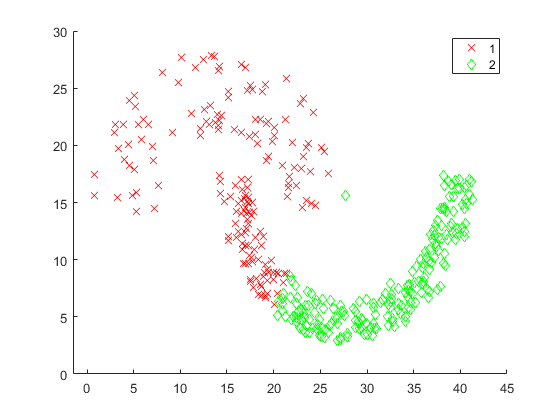}
                  \caption{$k$-means}
          \end{subfigure}%
          \begin{subfigure}[b]{0.5\textwidth}
                  \centering
                  \includegraphics[width=\textwidth]{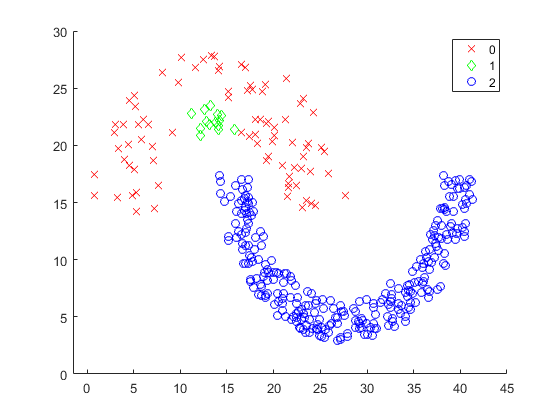}
                  \caption{DBSCAN}
          \end{subfigure}
                    \centering
          \begin{subfigure}[b]{0.5\textwidth}
                  \centering
                  \includegraphics[width=\textwidth]{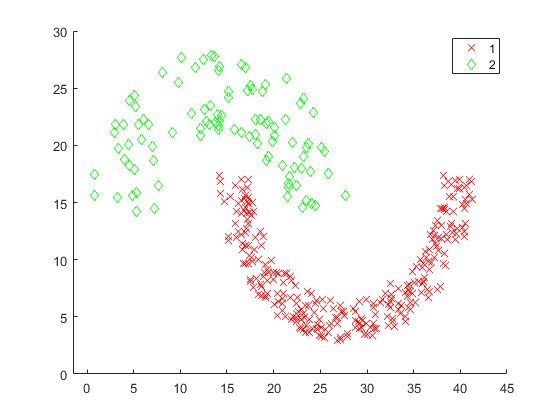}
                  \caption{ISDBSCAN}
          \end{subfigure}%
          \begin{subfigure}[b]{0.5\textwidth}
                  \centering
                  \includegraphics[width=\textwidth]{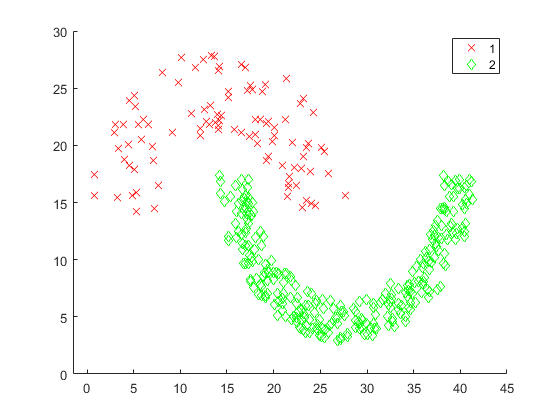}
                  \caption{DBSCRN}
          \end{subfigure}
    \end{subfigure}
\end{figure}

Figure \ref{Fig:Pathbased_BestPossible} presents the best possible clusterings for each algorithm in the Pathbased data set. This data set contains three clusters of equal cardinality in close proximity. These are separated by uneven low density regions. The clustering task is particularly difficult in this data set because two of the clusters are nested inside the third one. Unfortunately, $k$-means cannot deal with this type of scenario. DBSCAN and ISDBSCAN seem to find noise entities where there should not be any. The clusterings for the Toy data set can be seen in Figure \ref{Fig:Toy_BestPossible}. This data set contains two half-moon clusters of different densities. In these we can see that ISDBSCAN and DBSCRN were the only to correctly recover the two clusters. 

\begin{figure}[!htb]
    \centering
    \caption{Maximum run-time for DBSCAN, ISDBSCAN, and DBSCRN.}
    \label{Fig:AlgorithmsMaxRuntime}
    \includegraphics[width=8cm]{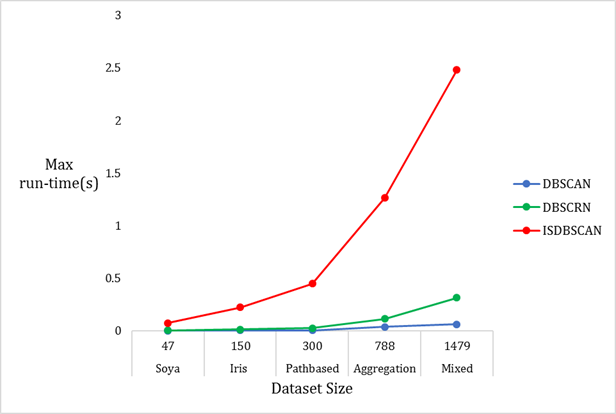}    
\end{figure}

Given the data sets we selected for our experiments it is hardly surprising that the density-based algorithms outperformed $k$-means in most cases. This result should not be interpreted as meaning that density-based algorithms tend to outperform distance-based algorithms. Before clustering what one ought to do is to define the objective of the clustering and then decide what method to use. Finally, in terms of conversion time we can see that DBSCAN is undoubtedly the fastest density-based algorithm we experiment with (see Figure \ref{Fig:AlgorithmsMaxRuntime}). However, DBSCAN has the worst cluster recovery and it is outperformed by ISDBSCAN and DBSCRN. DBSCRN outperforms ISDBSCAN in terms of cluster recovery and it is orders of magnitude faster than the latter. 

We find the results of our previous set of experiments very enlightening, but we feel we need to evaluate the algorithms in a realistic clustering scenario. We know DBSCRN has the best possible cluster recovery in most cases, but now we need to establish whether we can successfully estimate its parameters. With this in mind we ran a new set of experiments in which the parameters of each algorithm were those optimising the DBCV index. This is a truly unsupervised scenario. Table \ref{Tb:Results_UsingDBCV} presents the results in terms of cluster recovery. This time we decided not to run experiments with $k$-means because we have empirically demonstrated this is not well-aligned with the type of data sets we experiment with, and because DBCV was designed to be used by density-based algorithms. The experiments clearly demonstrate that in all cases DBSCRN is competitive or better than the other density-based algorithms we experiment with.

In our experiments we have shown that ISDBSCAN outperforms DBSCAN in terms of cluster recovery, and that DBSCRN outperforms both of them in the same measure. Table \ref{Tb:Speed} summarises the running time for each algorithm in seconds. This table includes the computational time required to run DBCV. We can clearly see DBSCAN is the fastest algorithm we experiment with. However, DBSCRN outperforms DBSCAN and ISDBSCAN in terms of cluster recovery, and it is orders of magnitude faster than the latter.
\begin{table*}[!htbp]
\caption{Experiments comparing DBSCAN, ISDBSCAN, and DBSCRN. The final clustering of each algorithm is that with the highest DBCV index.}
\centering
\label{Tb:Results_UsingDBCV}
\begin{tabular}{l l l l l l l l l l}
\toprule
\multirow{2}{*}{} & \multicolumn{3}{c}{DBSCAN} & \multicolumn{3}{c}{ISDBSCAN} & \multicolumn{2}{c}{DBSCRN}\\ 
\cmidrule(lr){2-4} \cmidrule(lr){5-7} \cmidrule(lr){8-9}
  & Mean & Std dev & Max & Mean & Std dev & Max  & Mean & Max  \\ \midrule
Aggregation & 0.95  & 0.01   & 0.97 &0.95  & 0.01   & 0.97  & -  & \textbf{0.99}  \\
Compound    & 0.75  & 0.00   & 0.75 &0.89  & 0.00   & 0.89    & -  & \textbf{0.96} \\
Pathbased   & 0.8   & 0.00   & 0.8 &0.55  & 0.02   & 0.6    & -  & \textbf{0.92} \\
Spiral      & \textbf{1.00}  & 0.00   & 1.00 & \textbf{1.00}  & 0.00   & 1.00   & -  & \textbf{1.00} \\
Mixed       & \textbf{1.00}  & 0.00   & 1.00 & \textbf{1.00}  & 0.00   & 1.00   & -  & \textbf{1.00} \\
Toy         & 0.36  & 0.00   & 0.36 & \textbf{1.00}  & 0.00   & 1.00    &  -  & \textbf{1.00} \\ 
Flame       & 0.88  & 0.01   & 0.9 & 0.92  & 0.01   & 0.94    & -  & \textbf{0.93} \\ 
R15         & \textbf{0.98}  & 0.00   & 0.98 & \textbf{0.98}  & 0.00   & 0.98    &  -  & 0.96 \\
Soya        & 0.98  & 0.00   & 1.00& 0.98  & 0.00   &1.00     & -  & \textbf{1.00} \\ 
Iris        & 0.36  & 0.01   & 0.37& 0.42  & 0.01   & 0.44     & -  & \textbf{0.45} \\ \hline \bottomrule
\end{tabular}
\end{table*}
\begin{table*}[!htbp]\small
\caption{Experiments comparing the run time in seconds of DBSCAN, ISDBSCAN, and DBSCRN. The below includes the time-elapse for DBCV.}
\centering
\label{Tb:Speed}
\begin{tabular}{l c c c c c c c c c c c c}
\toprule
\multirow{2}{*}{} & \multicolumn{4}{c}{DBSCAN} & \multicolumn{4}{c}{ISDBSCAN} & \multicolumn{4}{c}{DBSCRN}\\ 
\cmidrule(lr){2-5} \cmidrule(lr){6-9} \cmidrule(lr){10-13}
& Mean & Std dev & Max & Min & Mean & Std dev & Max & Min & Mean & Std dev & Max & Min  \\ \midrule
Aggregation&0.0237&0.0065&0.0374&0.0125&1.2481&0.0073&1.2655&1.2307&0.1131&0.0004&0.1149&0.1126\\
Compound&0.0071&0.0017&0.0101&0.0037&0.6043&0.0200&0.6819&0.5964&0.0490&0.0004&0.0508&0.0485\\
Flame&0.0032&0.0005&0.0065&0.0018&0.3458&0.0121&0.3709&0.3377&0.0194&0.0001&0.0201&0.0192\\
Iris&0.0018&0.0001&0.0036&0.0016&0.2221&0.0006&0.2247&0.2217&0.0130&0.0001&0.0135&0.0128\\
Mixed&0.0442&0.0037&0.0637&0.0349&2.4784&0.0019&2.4820&2.4759&0.3075&0.0053&0.3150&0.3034\\
Pathbased&0.0042&0.0010&0.0062&0.0022&0.4474&0.0003&0.4480&0.4467&0.0260&0.0001&0.0263&0.0258\\
R15&0.0177&0.0029&0.0217&0.0093&0.9157&0.0005&0.9167&0.9149&0.1948&0.0006&0.1986&0.1940\\
Soya&0.0005&0.0001&0.0012&0.0003&0.0747&0.0001&0.0750&0.0745&0.0042&0.0000&0.0044&0.0042\\
Spiral&0.0029&0.0004&0.0044&0.0025&0.2983&0.0012&0.3010&0.2966&0.0169&0.0001&0.0174&0.0168\\
Toy&0.0058&0.0014&0.0085&0.0032&0.5616&0.0006&0.5624&0.5605&0.0488&0.0002&0.0497&0.0485\\
\hline \bottomrule
\end{tabular}
\end{table*}
\section{Conclusion}
In this paper we have introduced a new density-based clustering algorithm, Density-based spatial clustering using reverse nearest neighbour (DBSCRN). We have run a number of experiments clearly showing our algorithm to outperform DBSCAN and ISDBSCAN in terms of cluster recovery. These experiments also established that we can indeed estimate a good parameter for DBSCRN which leads to better cluster recovery than that of other algorithms in a truly unsupervised scenario. Our experiments also show DBSCRN to be orders of magnitude faster than ISDBSCAN.

The experiments have also shown $k-$means not to perform well in most cases. Given the data sets we experiment with, this is hardly surprising. These results should not lead to conclusion that $k-$means is inferior to density-based algorithms, but rather that one should pay considerable attention when selecting a clustering algorithm. 

In our future research we intend to establish whether DBCV is indeed the best CVI to use in our case, and whether we can introduce the concept of feature weights to our method. These feature weights should model the degree of relevance of each feature in the data set.

\bibliographystyle{splncs_srt}
\bibliography{My_Bibliography}
\end{document}